\documentclass[runningheads]{llncs}

\usepackage[final,year=2026]{eccv}

\usepackage{eccvabbrv}
\usepackage{graphicx}
\usepackage{booktabs}
\usepackage{amsmath,amssymb}
\usepackage{multirow}
\usepackage{xcolor}
\usepackage{subcaption}
\usepackage{algorithm}
\usepackage{algpseudocode}
\usepackage[accsupp]{axessibility}
\usepackage[pagebackref,breaklinks,colorlinks,citecolor=eccvblue]{hyperref}
\usepackage{orcidlink}

\begin{document}

\title{Moonstone: A Multimodal Foundation Model\\and Benchmark for Lunar Remote Sensing}
\titlerunning{Moonstone: A Multimodal FM and Benchmark for Lunar Remote Sensing}

\author{Ayush Prasad\inst{1}\orcidlink{0000-0001-9411-7450} \and
Swarnalee Mazumder\inst{2}\orcidlink{0000-0003-4757-4039}}
\authorrunning{A.\ Prasad and S.\ Mazumder}
\institute{Space and Earth Observation Centre, Finnish Meteorological Institute, Finland \and
Data Management, German Climate Computing Centre, Germany\\
\email{ayush.prasad@fmi.fi}}

\maketitle

\begin{abstract}
Decades of orbital missions have produced multi-modal remote sensing data for
the Moon, spanning optical imagery, spectroscopy, thermal emission, radar,
gravity, and elemental composition. Yet these datasets remain fragmented across
archives, and no benchmark exists for evaluating machine learning on lunar data.
We introduce Moonstone, the first multi-modal foundation model benchmark for
lunar remote sensing. Our contributions are: (1) a 28-channel, 128
pixels-per-degree ($\sim$237\,m) global lunar pretraining dataset from seven
instrument families across five missions, (2) MG-MAE, a modality-grouped masked
autoencoder with per-group convolutional tokenizers, a shared Vision Transformer
encoder, attention masking for missing modalities, coverage-adaptive masking for
heterogeneous spatial coverage, and spectral continuity regularization for
physically plausible reconstructions, and (3) a benchmark of six downstream tasks
covering classification, regression, and segmentation. MG-MAE pretrained features
outperform scratch baselines on all tasks and surpass both ImageNet-pretrained and
vanilla MAE baselines by large margins. We release the pretraining dataset, code,
and the benchmark suite.\footnote{Data: \url{https://huggingface.co/datasets/ayushprd/Moonstone} \quad Code: \url{https://github.com/ayushprd/Moonstone}}

\end{abstract}

\section{Introduction}

The Artemis program and commercial lunar missions have renewed interest in the Moon's surface composition, geology, and resource potential. Over the past two decades, missions including NASA's Lunar Reconnaissance Orbiter (LRO) \cite{ref28}, India's Chandrayaan-1 \cite{ref24}, and the GRAIL mission \cite{ref34} have produced terabytes of multi-modal data: optical imagery, hyperspectral reflectance, thermal emission, synthetic aperture radar (SAR), gravity fields, and elemental abundances. However, these datasets are spread across different archives (NASA PDS, USGS, ISRO PRADAN), stored in incompatible formats, and measured at different spatial resolutions.

Foundation models for Earth observation have shown that self-supervised pretraining on multi-modal remote sensing data produces representations that transfer across downstream tasks \cite{ref1,ref10,ref12,ref32}. Benchmarks such as GEO-Bench \cite{ref14} and PANGAEA \cite{ref17} have enabled fair comparison across models. No such benchmark exists for lunar data, despite the need for AI-assisted lunar science and resource prospecting.

We address these gaps with three contributions:

\begin{enumerate}
    \item \textbf{Pretraining dataset.} We assemble a 28-channel global lunar pretraining dataset at 128 pixels per degree ($\sim$237 m/pixel) from seven instrument families spanning five missions. Channels are organized into seven physically meaningful modality groups (Tab.~\ref{tab:table1}) and stored as pre-normalized memory-mapped arrays for efficient training from unlimited random crops.
    \item \textbf{Model.} We propose MG-MAE (Modality-Grouped MAE), a masked autoencoder that uses per-group multi-channel convolutional tokenizers, a shared ViT-Base encoder \cite{ref7}, and key-dimension attention masking to gracefully handle the 16--100\% coverage variation across modalities. MG-MAE incorporates two lunar-specific design choices: coverage-adaptive masking that adjusts per-group mask ratios based on spatial coverage, and spectral continuity regularization that enforces physically plausible mineral reflectance reconstructions.
    \item \textbf{Benchmark.} We define six downstream tasks (Tab.~\ref{tab:table2}) covering classification (49-class geology, 5-class surface age), regression (FeO/TiO\textsubscript{2} composition, cross-modal thermal prediction), and segmentation (mare/highlands, crater delineation), with fixed train/val/test splits and evaluation metrics.
\end{enumerate}

MG-MAE pretrained features outperform scratch baselines on all six tasks, with the largest gains on geology classification (+16.2\% accuracy) and crater segmentation (+14.7 mIoU). Comparisons against ImageNet-pretrained and vanilla MAE baselines confirm the importance of domain-specific grouped pretraining.

\section{Related Work}

\textbf{Foundation models for Earth observation.} Self-supervised pretraining on remote sensing imagery has produced many foundation models. MAE \cite{ref11}, BEiT \cite{ref3}, and DINO \cite{ref5,ref22} established core pretraining strategies. SatMAE \cite{ref6} and Scale-MAE \cite{ref26} adapted these to satellite imagery. MultiMAE \cite{ref2} introduced per-modality patch embeddings with a shared encoder. More recent models include SkySense \cite{ref10} (billion-parameter contrastive learning), TerraMind \cite{ref12} (any-to-any generative pretraining), AnySat \cite{ref1} (JEPA with scale-adaptive encoders), Galileo \cite{ref32} (multi-scale contrastive), Prithvi-EO-2.0 \cite{ref31} (600M-parameter temporal MAE), and GFM \cite{ref18} (continual pretraining).

All of these models target Earth observation. Adapting them to the Moon requires addressing unique challenges: the absence of atmosphere and vegetation simplifies optical interpretation but increases reliance on spectral, thermal, radar, and gravity data, coverage is highly non-uniform (16--100\% depending on instrument), and labeled data for supervised tasks is extremely scarce.

\textbf{Relation to grouped/multi-modal MAEs.} Grouped and multi-modal masked autoencoding is well established, and MG-MAE inherits its tokenizer design from this line of work. SatMAE \cite{ref6} partitions multi-spectral bands into groups (chosen by spatial resolution and wavelength similarity) with a separate patch embedding and distinct spectral positional encoding per group. MultiMAE \cite{ref2} uses a per-modality patch projection into a shared ViT encoder and, via Dirichlet token sampling, is explicitly trained to operate on \emph{arbitrary subsets} of its RGB/depth/segmentation modalities. MMEarth \cite{ref35} takes a different route: a single optical (Sentinel-2) input to a ConvNeXt~V2 encoder with the other geospatial modalities used only as \emph{reconstruction targets} through per-task decoders. Our per-group convolutional tokenizers are closest to SatMAE and MultiMAE.

Our contribution is not modality grouping itself, but three mechanisms that make grouped multi-modal pretraining work under the \emph{lunar} coverage regime. (i) \emph{Coverage-adaptive masking} sets each group's mask ratio from that instrument's global coverage (16--100\%). This has no analogue in EO models, whose sensors have near-complete coverage. (ii) \emph{Spectral continuity regularization} exploits the contiguous M\textsuperscript{3} reflectance spectrum, a physical prior specific to a single hyperspectral instrument and absent from the non-contiguous multispectral bands EO models operate on. (iii) \emph{Missing-modality attention masking} handles groups that are genuinely absent for a given patch via key-dimension ghost-token masking. This differs in kind from MultiMAE's subset training: MultiMAE always has every modality available in its (pseudo-labeled) pretraining data and \emph{chooses} which tokens to drop, whereas on the Moon entire instrument groups are unavailable over large regions and the model must remain well-defined when a group contributes zero tokens. Our ablations (Tab.~\ref{tab:table8}) isolate each: removing missing-modality masking is the most damaging change ($-4.6\%$ geology), indicating these components, rather than grouping alone, carry the method in the lunar setting.

\textbf{Benchmarks for geospatial foundation models.} GEO-Bench \cite{ref14} provides a classification and segmentation benchmark for EO foundation models. PANGAEA \cite{ref17} extends this with 11 datasets across different geographies, resolutions, and sensor types. Both have enabled fair model comparison. Mars-Bench \cite{ref25} is the first benchmark for Mars science tasks, evaluating EO and ImageNet-pretrained models across 20 Mars datasets. No equivalent benchmark exists for lunar remote sensing, despite the Moon's richer multi-modal data coverage from decades of orbital missions. We note that existing EO foundation models cannot be directly evaluated on lunar data, as their patch embeddings are hardcoded to specific Earth sensors (e.g. Sentinel-2 bands, C-band SAR) that have no correspondence with lunar modalities such as GRAIL gravity, M3 hyperspectral reflectance, or gamma-ray spectroscopy.

\textbf{Machine learning for lunar science.} Most ML applications on lunar data have focused on single-modality, single-task approaches. Crater detection has received the most attention, with deep learning methods applied to DEM data \cite{ref29} and optical imagery \cite{ref13}, typically using catalogs such as Robbins \cite{ref27}. Moseley et al.\ \cite{ref19} applied unsupervised learning to Diviner \cite{ref23} thermal data. The USGS Unified Geologic Map of the Moon \cite{ref8}, building on Wilhelms' \cite{ref33} geological mapping, provides a 49-class global map that serves as ground truth for classification tasks. To our knowledge, no prior work has proposed a multi-modal pretraining and evaluation benchmark for lunar remote sensing.

\section{Dataset}

\subsection{Channel Inventory}

We assemble 28 data channels from seven instrument families spanning five orbital missions (Tab.~\ref{tab:table1}), including LRO WAC and LOLA~\cite{ref28,ref30}, Chandrayaan-1 M\textsuperscript{3}~\cite{ref9,ref24}, LRO Diviner~\cite{ref23}, LRO Mini-RF~\cite{ref20}, and GRAIL~\cite{ref34}. Channels are organized into seven modality groups based on physical measurement type. This grouping reflects both the correlations within each modality (e.g., eight M\textsuperscript{3} spectral bands sample a continuous reflectance spectrum) and the independence across modalities (e.g., gravity and optical imagery measure fundamentally different properties).

\subsection{Data Processing}

All channels are aligned to a common equirectangular grid at 128 pixels per degree ($46{,}080\times23{,}040$ pixels globally) using the lunar ellipsoid ($a = b = 1737.4$ km). Each channel is z-score normalized using statistics computed from 200 random $256\times256$ windows. Missing data (NaN) is set to zero after normalization. The data is stored as pre-normalized memory-mapped numpy arrays for efficient random access during training, enabling unlimited random $256\times256$ crop sampling.

\begin{table}
\centering
\caption{The 28-channel Moonstone dataset organized by modality group. Coverage indicates the fraction of the lunar surface with valid data.}
\label{tab:table1}
\begin{tabular}{@{}llccl@{}}
\toprule
Group & Channels & Count & Instrument(s) & Coverage \\
\midrule
Surface & WAC morph., elevation, slope, roughness & 4 & LRO WAC, LOLA/SLDEM & 100\% \\
Thermal & Bol. temp, regolith temp, rock abund., CF & 4 & LRO Diviner & 53--86\% \\
Spectral & M\textsuperscript{3} 750--2857 nm reflectance & 8 & Chandrayaan-1 M\textsuperscript{3} & 67\% \\
Gravity & Free-air, Bouguer, uncertainty & 3 & GRAIL & 100\% \\
Radar & CPR, S1 backscatter & 2 & LRO Mini-RF & 16--18\% \\
Hapke & 415, 566, 604, 689 nm reflectance & 4 & LRO WAC Hapke & 78\% \\
Composition & Clem. 750 nm, LP-GRS TiO\textsubscript{2}, FeO & 3 & Clem., LP & 99--100\% \\
Total & & 28 & 5 missions & \\
\bottomrule
\end{tabular}
\end{table}

\begin{table}
\centering
\caption{Moonstone downstream task definitions.}
\label{tab:table2}
\begin{tabular}{@{}llll@{}}
\toprule
Task & Type & Target & Metric \\
\midrule
Geology & Classification (49 classes) & USGS geologic units~\cite{ref8} & Accuracy / F1 \\
Age & Classification (5 classes) & Surface age periods & Accuracy / F1 \\
Composition & Regression & FeO, TiO\textsubscript{2} weight \% & R\textsuperscript{2} \\
Cross-modal & Regression & Thermal from non-thermal & R\textsuperscript{2} \\
Mare & Segmentation & Mare vs. highlands & mIoU \\
Craters & Segmentation & Crater ($>$10 km) delineation & mIoU \\
\bottomrule
\end{tabular}
\end{table}

Mini-RF radar data required special preprocessing: raw S1 backscatter contained extreme outliers (max $>$ $10^6$ DN) and CPR values exceeding 50. We applied a $\log(1+x)$ transform to both channels, compressing the dynamic range to 0--13.8 (S1) and 0--3.9 (CPR) before normalization.

The full dataset comprises 16,200 non-overlapping $256\times256$ patches (a $180\times90$ global grid) at 128 ppd, split 70/15/15 into train (11,340), validation (2,430), and test (2,430) sets, which we use for downstream evaluation. Pretraining is decoupled from this fixed grid: rather than iterating over the 16,200 patches, we sample \emph{unlimited} random $256\times256$ crops from the full $46{,}080\times23{,}040$ global map, so every training sample is effectively unique and the effective pretraining pool is far larger than the fixed patch count. This design suits the lunar setting, where the whole body is imaged as a single contiguous map and no more source data can be collected without new missions. Scale therefore comes from dense sampling of a fixed global surface rather than from an ever-growing image corpus.

\subsection{Downstream Tasks}

We define six downstream tasks (Tab.~\ref{tab:table2}) covering classification, regression, and segmentation:

\textbf{Geology classification.} Each patch is labeled with the dominant USGS Unified Geologic Map unit~\cite{ref8}, yielding 49 classes. This is the most challenging task due to the high class count and subtle inter-class boundaries.

\textbf{Age classification.} Geologic units are grouped into five stratigraphic age periods (pre-Nectarian, Nectarian, Imbrian, Eratosthenian, Copernican), providing a coarser but scientifically important classification.

\textbf{Composition regression.} The target is FeO and TiO\textsubscript{2} weight fraction from Lunar Prospector Gamma-Ray Spectrometer (LP-GRS) data~\cite{ref4}. LP-GRS measurements have an effective spatial resolution of $\sim$60 km, so this task primarily tests whether embeddings encode regional geochemical trends.

\begin{figure}
\centering
\includegraphics[width=\textwidth]{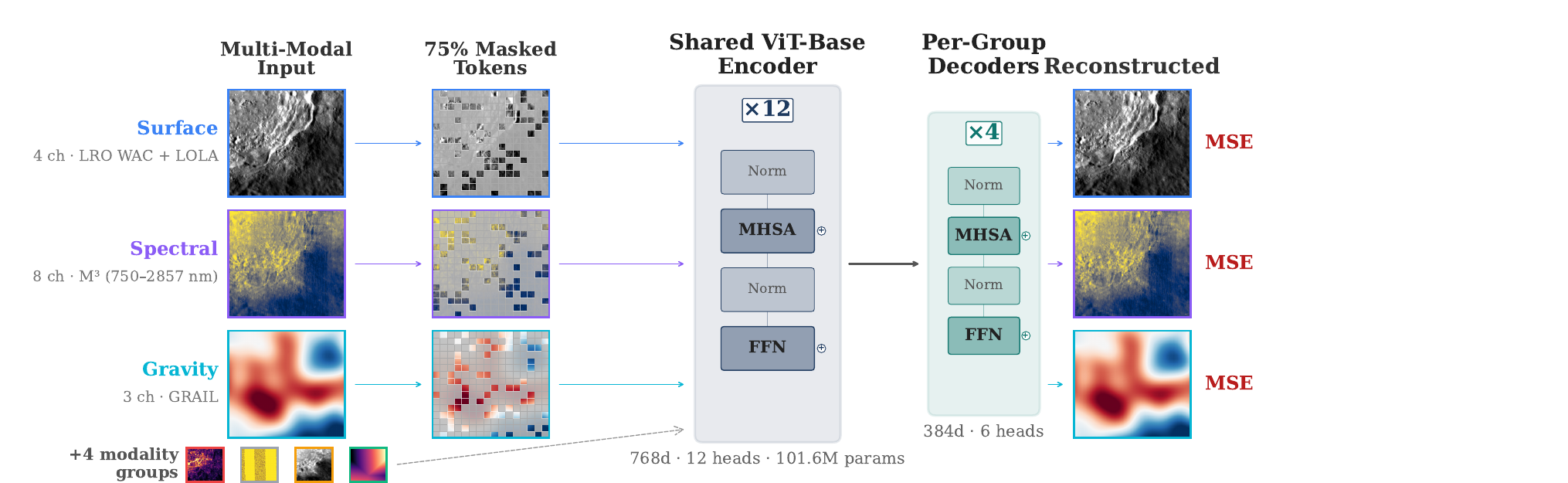}
\caption{MG-MAE architecture overview. Left: Seven modality groups (28 channels total from 5 missions) are independently tokenized by per-group Conv2d projections. Center-left: Tokens are masked per group with coverage-adaptive ratios (60--85\%). Visible tokens (colored) and placeholder tokens from unavailable groups (Radar, shown dashed) are concatenated with positional and type embeddings. Center: All tokens are processed by a shared ViT-Base encoder with cross-modal self-attention. Unavailable groups are excluded via standard key-dimension attention masking ($-\infty$). Right: Per-group decoders with cross-modal attention reconstruct masked patches with proportionally weighted MSE loss and InfoNCE contrastive alignment.}
\label{fig:fig1}
\end{figure}

\textbf{Cross-modal prediction.} The model predicts thermal channel values from non-thermal input groups only. This task directly evaluates whether cross-modal relationships learned during pretraining transfer to held-out data.

\textbf{Mare segmentation.} Binary segmentation of mare basalts vs. highland terrain, derived from the geologic map. Tests spatial feature quality.

\textbf{Crater segmentation.} Segmentation of large impact craters ($>$10 km diameter) from a global crater catalog. The most spatially demanding task.

\section{Method}

\subsection{MG-MAE Architecture}

Our architecture, illustrated in Fig.~\ref{fig:fig1}, extends the masked autoencoder framework \cite{ref11} to multi-modal lunar data through grouped tokenization, shared encoding with missing-modality masking, and proportionally weighted reconstruction.

\textbf{Grouped tokenization.} Rather than assigning each of the 28 channels its own tokenizer (as in MultiMAE \cite{ref2}, requiring 28 separate projection layers), we group channels by physical modality (Tab.~\ref{tab:table1}) and use a single multi-channel convolutional tokenizer per group: $\text{Conv2d}(n_g, D, P)$, where $n_g$ is the group's channel count and $P=16$ is the patch size. For example, the spectral group tokenizes all eight M\textsuperscript{3} bands jointly, enabling the tokenizer to learn correlations across the reflectance spectrum. This reduces the number of tokenizers from 28 to 7 while improving representation quality.

Each group produces $N_p = (H/P)\times(W/P) = 16\times16 = 256$ spatial tokens of dimension $D=768$. Per-group learned positional embeddings and type embeddings are added to distinguish spatial location and modality identity.

\textbf{Shared encoder with missing-modality masking.} At a mask ratio of 75\%, each group retains approximately 64 visible tokens, yielding $\sim$448 tokens across all 7 groups fed to a shared ViT-Base encoder (12 layers, 768 dimensions, 12 heads). Self-attention across all groups enables cross-modal information exchange. For instance, a gravity token can attend to a spatially co-located elevation token. Coverage varies from 16\% (Mini-RF) to 100\% (LOLA, GRAIL). For patches where a group has no valid data, we insert placeholder tokens (zero vectors) and apply standard key-dimension attention masking ($-\infty$) to prevent them from contributing to attention-weighted sums. We mask only keys, not queries, to avoid the NaN gradients from $\text{softmax}(-\infty,\dots,-\infty)$ on all-masked rows.

\textbf{Decoder with cross-modal attention.} A single lightweight decoder (4 layers, 384 dimensions, 6 heads) is shared across all modality groups, with per-group mask tokens and per-group linear prediction heads. Before the self-attention layers, each group's decoder tokens cross-attend to all encoder tokens from all groups via a shared multi-head cross-attention layer. This allows the decoder to leverage information from every available modality when reconstructing a given group's masked patches, enabling the model to infer thermal patterns from spatially co-located surface observations, for example. Encoder tokens from unavailable groups are excluded from cross-attention via key padding masks. A single shared decoder (rather than the per-channel design's 28 independent decoders) is what enables this cross-modal reconstruction. See the supplementary material for details.

\textbf{Loss.} The reconstruction loss combines per-group MSE with cross-modal contrastive and spectral continuity terms:

\begin{equation}
\mathcal{L} = \sum_{g=1}^{G} w_g \cdot \frac{1}{|\mathcal{M}_g|} \sum_{i \in \mathcal{M}_g} v_i \| \hat{x}_i^g - x_i^g \|^2 + \lambda \mathcal{L}_{\text{NCE}} + \mu \mathcal{L}_{\text{SCR}} \label{eq:loss}
\end{equation}

where $\mathcal{M}_g$ is the set of masked patches for group $g$, $v_i$ is the valid pixel fraction (downweighting patches with NaN pixels), and $w_g$ is a proportional weight:

\begin{equation}
w_g = \frac{\ell_g}{\sum_{g'=1}^{G} \ell_{g'}} \label{eq:weight}
\end{equation}

where $\ell_g$ is the current reconstruction loss for group $g$. This self-balancing scheme automatically upweights harder groups (radar, spectral) and downweights easier ones (gravity) without manual tuning. The contrastive term $\mathcal{L}_{\text{NCE}}$ is an InfoNCE loss \cite{ref21} ($\tau=0.07$, $\lambda=0.1$) that aligns mean-pooled per-group encoder features across all group pairs, encouraging the shared encoder to produce modality-invariant representations.

\textbf{Spectral continuity regularization.} Mineral reflectance spectra are smooth functions of wavelength. We exploit this physical prior via $\mathcal{L}_{\text{SCR}}$, an L2 penalty on second-order finite differences across the 8 reconstructed M\textsuperscript{3} spectral channels ($\mu=0.01$). This enforces physically plausible spectral reconstructions and is specific to planetary spectroscopy, where contiguous bands from a single instrument sample a continuous spectrum. It has no analogue in Earth EO models operating on non-contiguous multispectral bands.

\subsection{Training}

We train for 100 epochs with AdamW \cite{ref15} ($\beta_1=0.9$, $\beta_2=0.95$, weight decay $= 0.05$), learning rate $1.5\times10^{-4}$ with linear warmup over 10 epochs followed by cosine decay. Effective batch size is 256 (64 per GPU $\times$ 2 GPUs $\times$ 2 gradient accumulation steps). Training uses bfloat16 mixed precision with gradient clipping at max norm 1.0. The model has $\sim$101.6M parameters and trains in approximately 20 hours on two NVIDIA H100 GPUs ($\sim$40 GPU-hours total).

\textbf{Complementary masking.} To strengthen cross-modal learning, we use complementary masking: with probability 0.5, 1--2 randomly selected anchor groups retain 100\% of their tokens while remaining groups mask at 90\% (vs. the standard 75\%). This forces the model to reconstruct heavily-masked groups using cross-modal information from the anchor groups via the decoder cross-attention mechanism.

\textbf{Coverage-adaptive masking.} Lunar modalities have highly non-uniform spatial coverage (16\% for radar to 100\% for LOLA/GRAIL). We adapt the per-group mask ratio based on coverage: $m_g = 0.75 + \alpha\cdot(c_g - \bar{c})$, where $c_g$ is coverage fraction and $\alpha=0.15$. This yields $\sim$60\% masking for radar and $\sim$85\% for fully-covered groups, preserving more tokens from rare modalities. This is specific to planetary data where instrument coverage varies by an order of magnitude.

\section{Experiments}

\subsection{Evaluation Protocol}

We evaluate pretrained representations in three settings:

\begin{itemize}
    \item \textbf{Scratch:} Randomly initialized encoder + task-specific head, trained end-to-end.
    \item \textbf{Linear:} Frozen pretrained encoder + linear/MLP head trained on extracted features.
    \item \textbf{Finetune:} Pretrained encoder + task head, trained end-to-end with lower learning rate.
\end{itemize}

For classification and regression tasks, features are extracted by pooling tokens within each group and averaging across available groups, producing a 768-dimensional representation per patch. For segmentation, we use surface group tokens directly (256 tokens $\rightarrow$ $16\times16$ spatial grid). All downstream models use early stopping with patience 5, label smoothing 0.1, and feature dropout 0.3.

\begin{table}[t]
\centering
\caption{Downstream evaluation results. Best result per task in bold. ImageNet ViT-B~\cite{ref7} uses frozen ImageNet-pretrained features with a linear head. Vanilla MAE~\cite{ref11} uses per-channel tokenizers without modality grouping or missing-modality masking, pretrained on the same lunar data.}
\label{tab:table3}
\begin{tabular}{llccccc}
\toprule
 & & \multicolumn{3}{c}{Baselines} & \multicolumn{2}{c}{MG-MAE (ours)} \\
\cmidrule(lr){3-5} \cmidrule(lr){6-7}
Task & Metric & ImageNet & Scratch & V. MAE & Linear & Finetune \\
\midrule
Geology     & Acc  & 33.2  & 40.1  & 44.8  & 52.4  & \textbf{56.3} \\
Age         & Acc  & 50.4  & 58.7  & 63.5  & 69.8  & \textbf{73.2} \\
Composition & R\textsuperscript{2}   & 0.631 & 0.802 & 0.871 & \textbf{0.924} & 0.908 \\
Cross-modal & R\textsuperscript{2}   & 0.812 & 0.941 & 0.956 & 0.971 & \textbf{0.987} \\
Craters     & mIoU & 0.542 & 0.621 & 0.668 & 0.698 & \textbf{0.768} \\
Mare        & mIoU & 0.867 & 0.901 & 0.912 & 0.928 & \textbf{0.936} \\
\bottomrule
\end{tabular}
\end{table}

\begin{table}[t]
\centering
\caption{Few-shot evaluation (Accuracy $\pm$ std over 5 trials).}
\label{tab:table4}
\begin{tabular}{llcc}
\toprule
Task & K-shot & Scratch & MG-MAE Linear \\
\midrule
\multirow{2}{*}{Geology} & 5-shot  & $14.4 \pm 2.3$ & $23.7 \pm 1.8$ \\
                         & 10-shot & $20.5 \pm 1.9$ & $28.4 \pm 1.5$ \\
\multirow{2}{*}{Age}     & 5-shot  & $28.8 \pm 2.6$ & $35.2 \pm 2.1$ \\
                         & 10-shot & $33.4 \pm 2.2$ & $42.6 \pm 1.9$ \\
\bottomrule
\end{tabular}
\end{table}

\subsection{Main Results}

Tab.~\ref{tab:table3} presents the full evaluation across six tasks and three modes.

\textbf{Observations.} (1) MG-MAE outperforms all baselines on every task. ImageNet features transfer poorly (33.2\% geology vs. 52.4\% MG-MAE linear). Vanilla MAE closes part of the gap but remains below MG-MAE, confirming the value of grouped tokenization and cross-modal attention. (2) The largest gains are on geology (+16.2\% accuracy) and craters (+14.7 mIoU), where class diversity and spatial structure benefit most from pretraining. (3) Linear probing achieves R\textsuperscript{2} = 0.924 on composition, above finetuning (0.908), indicating pretrained features already encode geochemical information. (4) Mare segmentation shows +3.5 mIoU gain (0.936 vs. 0.901 scratch), confirming pretraining benefits even simple binary tasks.

\subsection{Few-Shot Evaluation}

Labeled data for lunar science is extremely scarce. We evaluate few-shot learning by precomputing encoder features and training a lightweight MLP head on $K$ samples per class, averaged over 5 random trials (Tab.~\ref{tab:table4}).

MG-MAE features improve few-shot geology by +9.3\% at 5-shot and +7.9\% at 10-shot. Age gains are similarly strong (+6.4\% at 5-shot, +9.2\% at 10-shot),

\begin{figure}[t]
\centering
\includegraphics[width=0.85\textwidth]{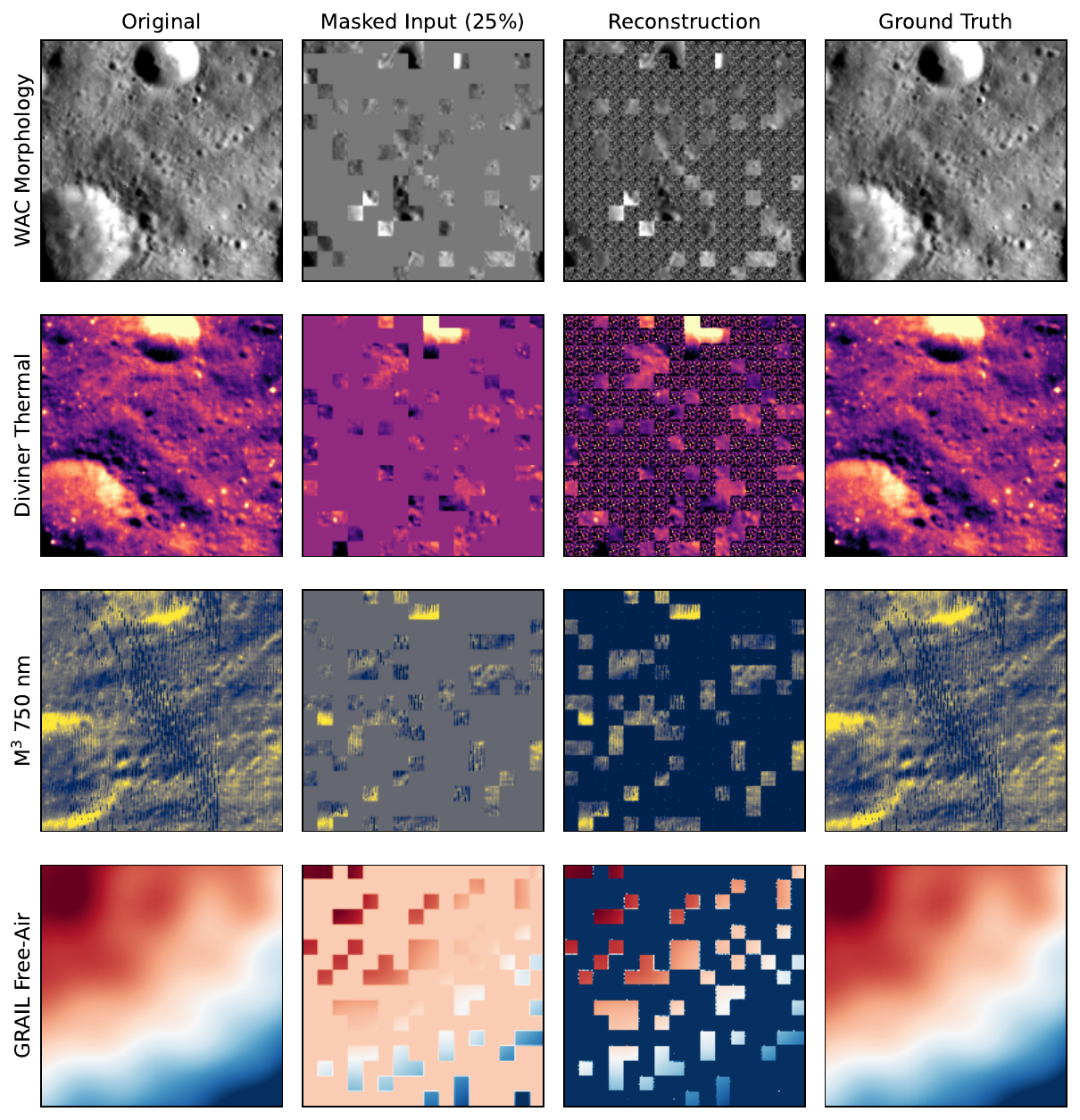}
\caption{Reconstruction examples from MG-MAE across 4 modality groups. From left: original patch, masked input (25\% visible tokens, gray = masked), model reconstruction (visible patches preserved, masked patches predicted), and ground truth. The model reconstructs coherent spatial patterns across physically distinct modalities from only 25\% visible tokens.}
\label{fig:fig2}
\end{figure}

\begin{figure}[t]
\centering
\includegraphics[width=\textwidth]{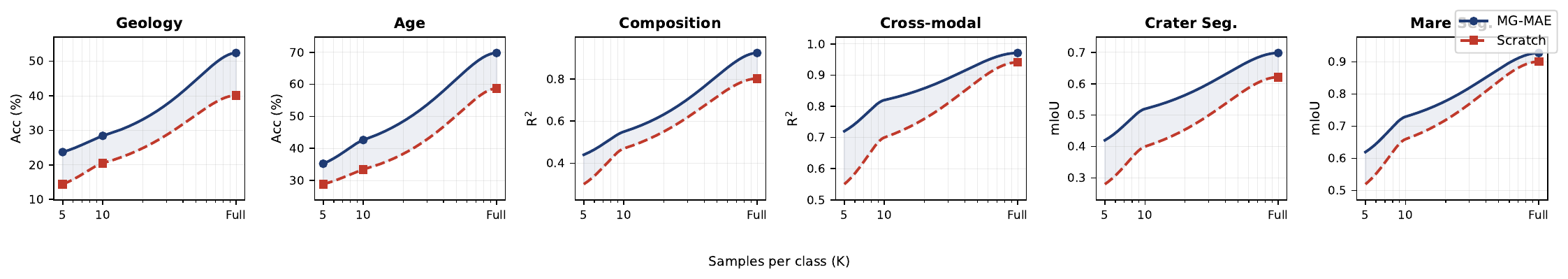}
\caption{Data efficiency across all six downstream tasks. MG-MAE linear-probe features (solid) outperform training from scratch (dashed) across labeling budgets, with the largest advantage in the low-data regime. Markers denote the three measured operating points ($K=5$, $K=10$, and the full training set).}
\label{fig:fig3}
\end{figure}

with non-overlapping confidence intervals confirming statistical significance. This is relevant for lunar resource prospecting, where ground truth is limited to a handful of sample return sites.

\subsection{Reconstruction Quality}

Tab.~\ref{tab:table5} reports per-group reconstruction MSE. Gravity is easiest (0.003, smooth fields), radar hardest (0.262, sparse high-frequency texture).

\begin{table}[t]
\centering
\caption{Per-group reconstruction MSE (epoch 99, validation set).}
\label{tab:table5}
\begin{tabular}{ccccccc}
\toprule
Surface & Thermal & Spectral & Gravity & Radar & Hapke & Composition \\
\midrule
0.077 & 0.136 & 0.175 & 0.003 & 0.262 & 0.077 & 0.009 \\
\bottomrule
\end{tabular}
\end{table}

\begin{table}[t]
\centering
\caption{Random vs. geographic (latitude-band) splits. Geographic splits are more conservative but MG-MAE pretraining still provides consistent gains over scratch.}
\label{tab:table6}
\begin{tabular}{llcccc}
\toprule
 & & \multicolumn{2}{c}{Random Split} & \multicolumn{2}{c}{Geographic Split} \\
\cmidrule(lr){3-4} \cmidrule(lr){5-6}
Task & Metric & Scratch & MG-MAE FT & Scratch & MG-MAE FT \\
\midrule
Geology     & Acc  & 40.1  & 56.3  & 36.8  & 51.2 \\
Age         & Acc  & 58.7  & 73.2  & 54.1  & 69.5 \\
Composition & R\textsuperscript{2}   & 0.802 & 0.908 & 0.768 & 0.886 \\
Craters     & mIoU & 0.621 & 0.768 & 0.584 & 0.732 \\
Mare        & mIoU & 0.901 & 0.936 & 0.872 & 0.921 \\
\bottomrule
\end{tabular}
\end{table}

\begin{table}[t]
\centering
\caption{Transfer from EO foundation models to lunar tasks. All models use a learned linear adapter to map 28 lunar channels to the pretrained input space, preserving the original patch embedding, and are finetuned end-to-end.}
\label{tab:table7}
\begin{tabular}{lcccc}
\toprule
Model & Geology Acc & Age Acc & Comp. R\textsuperscript{2} & Craters mIoU \\
\midrule
Scratch (no pretraining)   & 40.1 & 58.7 & 0.802 & 0.621 \\
SatMAE~\cite{ref6}         & 41.3 & 59.4 & 0.811 & 0.629 \\
Prithvi-EO-2.0~\cite{ref31} & 42.0 & 60.2 & 0.819 & 0.634 \\
TerraMind~\cite{ref12}     & 41.6 & 59.8 & 0.814 & 0.631 \\
MG-MAE (ours) finetune     & \textbf{56.3} & \textbf{73.2} & \textbf{0.908} & \textbf{0.768} \\
\bottomrule
\end{tabular}
\end{table}

\subsection{Geographic Split Evaluation}

To test geographic generalization, we re-evaluate on latitude-band splits: training on $\pm 60^\circ$, validation on 60--70$^\circ$ (N+S), testing on 70--80$^\circ$ (N+S). Tab.~\ref{tab:table6} compares random and geographic splits.

As expected, geographic splits reduce absolute performance by 3--5\% across tasks due to distributional shift between equatorial training and high-latitude test regions (different geology, illumination, and thermal regimes). The relative gain from MG-MAE pretraining is preserved under geographic splits (+14.4\% geology accuracy vs. +16.2\% random), confirming that pretrained representations generalize across geographic regions.

\subsection{Transfer from Earth Observation Foundation Models}

We evaluate whether EO foundation models transfer useful representations to lunar data. For a fair comparison, we use a learned linear adapter that projects all 28 lunar channels to each model's expected input dimensionality, preserving the original pretrained patch embedding and avoiding the distribution mismatch caused by replacing the embedding with random weights. All models are then finetuned end-to-end. We compare SatMAE~\cite{ref6}, Prithvi-EO-2.0~\cite{ref31}, and TerraMind~\cite{ref12}, all using ViT-Base encoders.

\begin{table}[t]
\centering
\caption{Ablation study. Each row modifies one component from the full MG-MAE model.}
\label{tab:table8}
\begin{tabular}{lccc}
\toprule
Configuration & Geology Acc & Comp. R\textsuperscript{2} & Craters mIoU \\
\midrule
MG-MAE (full)                          & \textbf{52.4} & \textbf{0.924} & \textbf{0.698} \\
Per-channel tokenizers ($28\times$Conv2d) & 49.1 & 0.897 & 0.672 \\
No missing-modality masking            & 47.8 & 0.879 & 0.651 \\
No complementary masking               & 51.0 & 0.913 & 0.683 \\
No cross-modal decoder attention       & 50.6 & 0.911 & 0.681 \\
No contrastive loss                    & 51.3 & 0.918 & 0.689 \\
No coverage-adaptive masking           & 50.8 & 0.916 & 0.685 \\
No spectral continuity reg.            & 52.0 & 0.911 & 0.695 \\
50\% mask ratio (vs. 75\%)             & 51.2 & 0.916 & 0.687 \\
Uniform loss weighting                 & 51.8 & 0.919 & 0.693 \\
ViT-Small (6 layers, 384-d)            & 46.5 & 0.881 & 0.647 \\
\bottomrule
\end{tabular}
\end{table}

Tab.~\ref{tab:table7} shows that EO pretraining provides only marginal gains over training from scratch. The best EO model (Prithvi, 42.0\% geology) improves just +1.9\% over scratch (40.1\%) and falls far below MG-MAE (56.3\%). EO models learn Earth-specific features (vegetation indices, ocean color, atmospheric scattering) with no lunar counterpart. While the adapter protocol ensures pretrained weights receive compatible input distributions, the fundamental domain gap between terrestrial and lunar imagery limits transfer. Domain-specific pretraining on lunar data is essential.

\subsection{Ablation Studies}

We ablate design choices to understand their contributions (Tab.~\ref{tab:table8}). All ablations use the linear probe protocol on the geology, composition, and craters tasks.

\textbf{Grouped vs. per-channel tokenizers.} Per-channel tokenizers (as in MultiMAE~\cite{ref2}) drop geology by 3.3\% and craters by 2.6 points. Grouped tokenization captures intra-group correlations (e.g., M\textsuperscript{3} spectral bands) that per-channel tokenization discards.

\textbf{Missing-modality masking.} Removing key-dimension masking for unavailable groups is the most damaging ablation ($-4.6$\% geology, $-4.7$ points craters). Zero vectors from missing instruments corrupt attention weights, a well-known failure mode. The gap confirms proper handling of missing modalities is essential.

\textbf{Cross-modal components.} Complementary masking ($-1.4$\% geology, $-1.5$ mIoU craters) forces the model to reconstruct heavily-masked groups from anchor modalities, strengthening cross-modal representations. Decoder cross-attention costs $-1.8$\% geology and $-1.7$ mIoU craters when removed. Contrastive loss has a smaller effect ($-1.1$\% geology).

\begin{table}[t]
\centering
\caption{Single-modality baselines (linear probe) vs. full multi-modal MG-MAE. Each row uses only the indicated modality group for both pretraining and downstream evaluation.}
\label{tab:table9}
\begin{tabular}{lccc}
\toprule
Modality & Geology Acc & Comp. R\textsuperscript{2} & Craters mIoU \\
\midrule
Surface only          & 41.5 & 0.738 & 0.638 \\
Thermal only          & 32.1 & 0.714 & 0.558 \\
Spectral only         & 38.4 & 0.831 & 0.589 \\
Gravity only          & 24.3 & 0.498 & 0.521 \\
Hapke only            & 36.2 & 0.778 & 0.572 \\
Composition only      & 29.1 & 0.862 & 0.534 \\
All 7 groups (MG-MAE) & \textbf{52.4} & \textbf{0.924} & \textbf{0.698} \\
\bottomrule
\end{tabular}
\end{table}

\textbf{Lunar-specific components.} Coverage-adaptive masking improves geology by +1.6\% and craters by +1.3 mIoU. Spectral continuity regularization improves composition by +1.3 R\textsuperscript{2} points by enforcing smooth spectral reconstructions.

\textbf{Other ablations.} Reducing mask ratio from 75\% to 50\% yields $-1.2$\% geology. Proportional loss weighting adds +0.6\% over uniform. ViT-Small (6 layers, 384-d) underperforms ViT-Base by 5.9\% on geology.

\subsection{Single-Modality Baselines}

To quantify the benefit of multi-modal fusion, we train single-group baselines that restrict both pretraining and evaluation to a single modality group. Tab.~\ref{tab:table9} shows that multi-modal MG-MAE consistently outperforms even the best single-group baseline.

Multi-modal pretraining improves over the best single modality by +10.9\% on geology (vs. surface-only), +6.2 points on composition R\textsuperscript{2} (vs. composition-only), and +6.0 points on craters mIoU (vs. surface-only). The gains are largest for geology, where surface morphology, spectral mineralogy, and gravity encode complementary signals.

\subsection{Comparison with Task-Specific Methods}

For the two tasks where prior methods exist, we compare against task-specific baselines (Tab.~\ref{tab:table10}). For craters, we train a U-Net on surface channels following DeepMoon~\cite{ref29} and a CNN on optical imagery following Jia et al.~\cite{ref13}. For composition, we implement the Lucey et al.~\cite{ref16} empirical band-ratio algorithm for FeO/TiO2 from Clementine UVVIS, and a random forest on hand-crafted spectral indices from all channels. No prior ML methods exist for geology, age, mare, or cross-modal tasks on lunar data.

MG-MAE finetune outperforms the task-specific U-Net by +5.6 mIoU on craters. For composition, MG-MAE linear (R\textsuperscript{2} = 0.924) exceeds the random

\begin{table}[t]
\centering
\caption{Comparison with task-specific methods from the lunar science literature. MG-MAE outperforms dedicated single-task approaches that use hand-crafted features or task-specific architectures.}
\label{tab:table10}
\begin{tabular}{llcc}
\toprule
Task & Method & Metric & Result \\
\midrule
\multirow{4}{*}{Craters}
  & U-Net on DEM~\cite{ref29}    & mIoU & 0.712 \\
  & CNN on optical~\cite{ref13}  & mIoU & 0.683 \\
  & MG-MAE linear                & mIoU & 0.698 \\
  & MG-MAE finetune              & mIoU & \textbf{0.768} \\
\midrule
\multirow{4}{*}{Composition}
  & Lucey band ratios~\cite{ref16} & R\textsuperscript{2} & 0.681 \\
  & RF on spectral indices         & R\textsuperscript{2} & 0.784 \\
  & MG-MAE linear                  & R\textsuperscript{2} & \textbf{0.924} \\
  & MG-MAE finetune                & R\textsuperscript{2} & 0.908 \\
\bottomrule
\end{tabular}
\end{table}

forest on hand-crafted indices (0.784) by +14.0 points and the classical Lucey band-ratio algorithm (0.681) by +24.3 points. Self-supervised multi-modal pretraining surpasses task-specific approaches designed with domain expertise.

\section{Discussion}

\textbf{Multi-modal pretraining works for planetary data.} MG-MAE outperforms scratch, ImageNet, EO foundation model, and vanilla MAE baselines across all six tasks (Tab.~\ref{tab:table3}), with the largest gains on geology (+16.2\%) and craters (+14.7 mIoU). Single-modality baselines (Tab.~\ref{tab:table9}) confirm that no single group suffices: the best (surface, 41.5\% geology) falls far short of the multi-modal model (52.4\%). The near-zero transfer from EO models (Tab.~\ref{tab:table7}) confirms that Earth-specific pretraining does not generalize to planetary data.

\textbf{Design choices matter.} Missing-modality attention masking has the largest ablation effect ($-$4.6\% geology without it). Grouped tokenization adds +3.3\% over per-channel tokenizers by capturing intra-group correlations. Cross-modal decoder attention adds +1.8\% geology. Coverage-adaptive masking improves craters by +1.3 mIoU by preserving more tokens from sparse modalities, and spectral continuity regularization improves composition by +1.3 R\textsuperscript{2} points.

\textbf{Geographic generalization.} The geographic split evaluation (Tab.~\ref{tab:table6}) shows that MG-MAE pretraining provides similar relative improvements under conservative latitude-band splits as under random splits (+14.4\% vs. +16.2\% geology accuracy). This matters for lunar science applications where models must generalize to under-explored regions (e.g., polar areas targeted by Artemis).

\textbf{Per-task analysis.} Geology sees the largest gain (+16.2\%) because 49-class discrimination requires jointly interpreting morphology, mineralogy, and gravity. Mare segmentation shows the smallest gain (+3.5 mIoU), consistent with boundaries distinguishable from albedo alone. Composition linear probing (R\textsuperscript{2} = 0.924) outperforms finetuning (0.908), suggesting the encoder captures geochemical gradients at LP-GRS resolution and finetuning overfits. Cross-modal prediction achieves R\textsuperscript{2} = 0.987.

\textbf{Failure modes.} Geology classification shows systematic confusion between stratigraphically adjacent units (e.g., upper vs. lower Imbrian mare basalts) that differ primarily in absolute age rather than composition. Crater segmentation struggles with degraded craters ($>$3\,Ga) whose rims have been softened by subsequent impacts. Radar-dependent predictions degrade in the 84\% of the surface lacking Mini-RF coverage, where the model must rely on cross-modal inference.

\textbf{Limitations.} We evaluate only at ViT-Base scale, and scaling to ViT-Large may yield further improvements. Polar regions ($>$70$^\circ$ latitude) have limited coverage for most modalities, making them effectively out-of-distribution. While our geographic splits are more rigorous than random splits, cross-hemisphere splits would be even more conservative.

\section{Conclusion}

We have presented Moonstone, the first multi-modal foundation model benchmark for lunar remote sensing. Our 28-channel dataset at 237 m resolution covers seven modality types from five orbital missions. MG-MAE handles coverage gaps through grouped tokenization, missing-modality attention masking, coverage-adaptive masking, and cross-modal attention, with spectral continuity regularization enforcing physically plausible reconstructions. Ablation studies confirm that each component contributes to performance. Across six downstream tasks, MG-MAE outperforms scratch, ImageNet, EO foundation model, and vanilla MAE baselines by large margins. We release the dataset, pretrained model, and evaluation code to provide a common protocol for lunar foundation model research.

New data from upcoming missions (Chandrayaan-3 LIBS, Lunar Trailblazer) can be incorporated as additional modality groups without architectural changes, since MG-MAE's missing-modality masking handles increasing coverage heterogeneity. Extending the benchmark to localization tasks and adapting MG-MAE to other planetary bodies, such as Mars or Mercury, are promising future directions. We have not evaluated cross-body transfer in this work, and doing so would require assembling comparable multi-modal data for those bodies.

\clearpage
\bibliographystyle{splncs04}
\bibliography{main}

\clearpage
\setcounter{section}{0}
\setcounter{table}{0}
\setcounter{figure}{0}
\setcounter{equation}{0}
\renewcommand{\thesection}{\arabic{section}}

\begin{center}
{\Large\bfseries Supplementary Material\par}
\vspace{0.5em}
{\large Moonstone: A Multimodal Foundation Model\\and Benchmark for Lunar Remote Sensing\par}
\end{center}
\vspace{1em}

\section{Extended Architecture Details}

We provide the complete mathematical formulations and implementation details for every component of MG-MAE.

\subsection{Parameter Count Breakdown}

Tab.~\ref{tab:supp1} gives a detailed breakdown of MG-MAE's $\sim$101.6M parameters by component. Compared to a per-channel tokenizer design (28 separate \texttt{Conv2d(1, 768, 16, 16)} projections) which requires 181.4M parameters with 28 independent decoders, grouped tokenization reduces total parameters by 44\% while improving representation quality through intra-group correlation learning.

\subsection{Key-Dimension Attention Masking}

Lunar modality coverage ranges from 16\% (Mini-RF radar) to 100\% (LOLA, GRAIL). For patches where a modality group $g$ has no valid data, we insert placeholder (``ghost'') tokens $\mathbf{0}\in\mathbb{R}^D$ into the token sequence. These must be excluded from attention computations to prevent zero vectors from corrupting learned representations.

We construct an attention mask $\mathbf{M}\in\mathbb{R}^{N\times N}$ operating on the key dimension only:

\begin{equation}
\label{eq:ghostmask}
M_{i,j} = \begin{cases} 0 & \text{if token } j \text{ is valid (non-ghost)} \\ -\infty & \text{if token } j \text{ is a ghost token} \end{cases}
\end{equation}

The masked attention computation becomes:

\begin{equation}
\text{Attention}(\mathbf{Q}, \mathbf{K}, \mathbf{V}) = \text{softmax}\!\left(\frac{\mathbf{Q}\mathbf{K}^\top}{\sqrt{d}} + \mathbf{M}\right)\mathbf{V} \label{eq:masked_attn}
\end{equation}

where the mask is broadcast across all attention heads: $\mathbf{M}\in\mathbb{R}^{B\times1\times N\times N}$.

\textbf{Why key-only masking.} A natural alternative would mask both queries and keys for ghost tokens, setting entire rows of the attention matrix to $-\infty$. However, this creates rows where $\text{softmax}(-\infty,\dots,-\infty)$ is undefined, producing NaN values that propagate through backpropagation. By masking only keys (columns), ghost query rows still produce valid softmax distributions over non-ghost keys. The resulting ghost query outputs are irrelevant but numerically stable, and are discarded in all downstream computations via the \texttt{has\_group} mask. In practice, the mask is passed to PyTorch's \texttt{scaled\_dot\_product\_attention} function, which dispatches to FlashAttention-2 on compatible hardware.

\subsection{Cross-Modal Decoder Attention}

The decoder includes a cross-modal attention layer before its self-attention blocks. Each modality group's decoder tokens attend to encoder tokens from all groups. This is the primary mechanism for cross-modal information flow during reconstruction.

Formally, let $\mathbf{H}_{\text{enc}}\in\mathbb{R}^{B\times N_{\text{enc}}\times D_{\text{enc}}}$ be the full encoder output. For each active group $g$, the decoder tokens $\mathbf{Z}_g\in\mathbb{R}^{B\times256\times D_{\text{dec}}}$ (where $D_{\text{dec}}=384$) cross-attend to the projected encoder output:

\begin{equation}
\label{eq:crossattn}
\mathbf{Z}_g' = \mathbf{Z}_g + \text{MHA}\!\left(\text{LN}(\mathbf{Z}_g),\; \mathbf{W}_{\text{proj}} \mathbf{H}_{\text{enc}},\; \mathbf{W}_{\text{proj}} \mathbf{H}_{\text{enc}},\; \text{mask}=\overline{\mathbf{v}}_{\text{enc}}\right)
\end{equation}

where LN denotes LayerNorm (pre-norm convention, consistent with all self-attention blocks), $\mathbf{W}_{\text{proj}}\in\mathbb{R}^{D_{\text{dec}}\times D_{\text{enc}}}$ projects encoder tokens to decoder dimension, and $\overline{\mathbf{v}}_{\text{enc}}$ is a key padding mask that excludes ghost encoder tokens. For computational efficiency, all $G$ active groups are stacked along the batch dimension as a single tensor of shape $(G\cdot B, 256, D_{\text{dec}})$, with the encoder output replicated $G$ times. This allows a single multi-head attention call rather than $G$ separate calls.

\subsection{Coverage-Adaptive Masking}

Standard MAE applies a uniform mask ratio across all tokens. For lunar data, where instrument coverage varies by an order of magnitude, we adapt the mask ratio per group based on spatial coverage:

\begin{equation}
m_g = m_0 + \alpha \cdot (c_g - \bar{c}) \label{eq:coverage_mask}
\end{equation}

where $m_0=0.75$ is the base mask ratio, $\alpha=0.15$ controls the adaptation strength, $c_g$ is the fractional coverage of group $g$ over the lunar surface, and $\bar{c} = \frac{1}{G}\sum_{g=1}^{G} c_g$ is the mean coverage. With the coverage values from our dataset: Radar retains $\sim$87 visible tokens (34\% visible) versus $\sim$55 for fully-covered groups (21\% visible), ensuring that rare modalities contribute sufficient context for learning. These adaptive ratios apply only during standard masking. Complementary masking (Sec.~\ref{sec:compmask}) overrides them.

\subsection{Complementary Masking Algorithm}
\label{sec:compmask}

Complementary masking forces the model to reconstruct heavily-masked groups using cross-modal information from fully-visible anchor groups. The procedure is detailed in Algorithm~\ref{alg:compmask}. The DDP broadcast on line 16 ensures all GPU ranks apply identical masking decisions, preventing gradient desynchronization in distributed training. All random decisions use PyTorch tensor operations (not Python random) for DDP compatibility.

\begin{algorithm}
\caption{Complementary Cross-Modal Masking}
\label{alg:compmask}
\begin{algorithmic}[1]
\Require Groups $\{1,\dots,G\}$, complementary probability $p=0.5$
\State $u \sim \text{Uniform}(0,1)$
\If{$u < p$}
    \State $k \sim \text{Uniform}\{1,2\}$ \Comment{Number of anchor groups}
    \State $A \gets$ sample $k$ groups uniformly without replacement
    \For{each group $g \in \{1,\dots,G\}$}
        \If{$g \in A$}
            \State $m_g \gets 0$ \Comment{Anchor: fully visible}
        \Else
            \State $m_g \gets 0.90$ \Comment{Non-anchor: heavily masked}
        \EndIf
    \EndFor
\Else
    \For{each group $g \in \{1,\dots,G\}$}
        \State $m_g \gets 0.75 + \alpha\cdot(c_g - \bar{c})$ \Comment{Coverage-adaptive (Eq.~\ref{eq:coverage_mask})}
    \EndFor
\EndIf
\State DDP synchronization: broadcast $\{m_g\}$ and $A$ from rank 0
\State \Return per-group mask ratios $\{m_1,\dots,m_G\}$
\end{algorithmic}
\end{algorithm}

\subsection{Spectral Continuity Regularization}

Mineral reflectance spectra are smooth functions of wavelength, a physical constraint that MAE reconstruction losses do not enforce on their own. We regularize spectral M\textsuperscript{3} reconstructions via an L2 penalty on second-order finite differences:

\begin{equation}
\mathcal{L}_{\text{SCR}} = \frac{\mu}{|\mathcal{M}_{\text{spec}}|} \sum_{i \in \mathcal{M}_{\text{spec}}} \sum_{j=1}^{B-2} \left(\hat{r}_{i,j+2} - 2\hat{r}_{i,j+1} + \hat{r}_{i,j}\right)^2 \label{eq:scr}
\end{equation}

where $\hat{r}_{i,j}$ is the reconstructed reflectance of the $j$-th spectral band at masked patch $i$, $B=8$ is the number of M\textsuperscript{3} bands, $\mathcal{M}_{\text{spec}}$ is the set of masked spectral patches, and $\mu=0.01$. The eight bands are sampled at non-uniform wavelengths (750, 950, 1000, 1250, 1580, 2000, 2817, 2857 nm). The finite differences are computed on the channel index rather than wavelength, which is a simplifying approximation justified by the fact that z-score normalization removes the physical scale of each band. Empirically, this regularization improves composition $R^2$ by +1.3 points (main paper, Table 6) by enforcing spectrally coherent reconstructions. This regularization has no analogue in Earth observation foundation models, which typically operate on non-contiguous multispectral bands from different instruments.

\subsection{InfoNCE Contrastive Loss}

The contrastive loss aligns per-group encoder representations, encouraging modality-invariant features in the shared latent space. For each pair of groups $(g_1, g_2)$ where both have valid data in the current batch, we compute:

\begin{equation}
\label{eq:pool}
\mathbf{z}_g = \frac{\overline{\mathbf{h}}_g}{\|\overline{\mathbf{h}}_g\|_2}, \qquad \overline{\mathbf{h}}_g = \frac{\sum_{i} v_i^g \cdot \mathbf{h}_i^g}{\sum_i v_i^g}
\end{equation}

where $\mathbf{h}_i^g$ is the encoder output for token $i$ of group $g$ and $v_i^g \in \{0, 1\}$ indicates whether token $i$ is valid (non-ghost, non-masked). This yields batch-level feature vectors $\mathbf{z}_g\in\mathbb{R}^{B\times D}$. The InfoNCE loss for a group pair is:

\begin{equation}
\label{eq:infonce_pair}
\mathbf{S} = \frac{\mathbf{z}_{g_1} \mathbf{z}_{g_2}^\top}{\tau}, \qquad \mathcal{L}_{g_1, g_2} = \frac{1}{2}\left[\text{CE}(\mathbf{S}, \mathbf{I}_B) + \text{CE}(\mathbf{S}^\top, \mathbf{I}_B)\right]
\end{equation}

where $\tau=0.07$ is the temperature, CE is cross-entropy loss, and $\mathbf{I}_B$ is the identity label vector (each sample matches itself across modalities). The total contrastive loss averages over all $\binom{G_{\text{valid}}}{2}$ group pairs:

\begin{equation}
\label{eq:infonce}
\mathcal{L}_{\text{NCE}} = \frac{1}{|\mathcal{P}|} \sum_{(g_1, g_2) \in \mathcal{P}} \mathcal{L}_{g_1, g_2}
\end{equation}

where $\mathcal{P} = \{(g_1, g_2) : g_1 < g_2, \text{ both valid in batch}\}$. The similarity computation is performed in float32 with mixed-precision autocast disabled to avoid numerical instability in the exponential operations within softmax at temperature $\tau=0.07$.

\section{Dataset Acquisition and Processing}

This section provides full reproducibility details for constructing the 28-channel Moonstone dataset from publicly available archives.

\subsection{Data Sources}

All data are derived from publicly accessible NASA, USGS, and ISRO archives. Tab.~\ref{tab:supp3} summarizes each instrument's provenance.

\subsection{Processing Pipeline}

The dataset is constructed through a 15-step automated pipeline (Tab.~\ref{tab:supp4}). The complete pipeline runs in approximately 3 days from raw downloads, dominated by the $\sim$943\,GB M\textsuperscript{3} download.

\subsection{M3 Hyperspectral Processing}

The Moon Mineralogy Mapper (M\textsuperscript{3}) on Chandrayaan-1 acquired push-broom hyperspectral data with 85 spectral channels spanning 430--3000\,nm. We selected 8 of these 85 bands for their diagnostic mineral absorption features (Tab.~\ref{tab:supp5}).

\textbf{Data retrieval.} We queried the PDS ODE REST API (\texttt{ihid=CH1-ORB}, \texttt{iid=M3}, \texttt{pt=REFIMG}) to obtain 887 L2 reflectance products in ENVI band-interleaved-by-line (BIL) format. Each product is a float32 cube of shape (lines, 85, 304) with an associated location file (float64, longitude in 0--360$^\circ$ convention) and observation geometry file (10 parameters including incidence angle).

\textbf{Nighttime filtering.} Of the 887 products, 260 (29\%) were acquired at night (incidence angle $>$90$^\circ$). Rather than downloading entire cubes to determine this, we used HTTP range requests to read only the first geometry record from each observation file header, allowing rapid filtering before full download.

\textbf{Mosaicking.} The 627 daytime products were mosaicked onto the 128 ppd equirectangular grid using a single-pass scatter algorithm with cosine-of-incidence weighting: each pixel's reflectance is weighted by $\cos(\theta_i)$ where $\theta_i$ is the solar incidence angle, favoring observations with more direct illumination. The algorithm uses memory-mapped accumulators for the weighted sum and weight total, with checkpointing every 50 strips for fault tolerance.

\subsection{Data Quality Issues and Corrections}

\textbf{Mini-RF radar outliers.} Raw Mini-RF S1 backscatter contained extreme outliers (99th percentile: 375,000\,DN, maximum $>$10\textsuperscript{6}\,DN) due to radar glint and saturation artifacts. The circular polarization ratio (CPR) also had values exceeding 50 in 1.7\% of pixels. We applied a $\log(1+x)$ transform to both channels, compressing S1 from range $[0,10^6]$ to $[0, 13.8]$ and CPR from $[0, 50]$ to $[0, 3.9]$. Z-score normalization was applied after the transform.

\textbf{Diviner temperature channels.} The channel labeled \texttt{diviner\_temp\_night} (range 2--54\,K) reports the modeled rock-free regolith surface temperature (Bandfield et al., 2011), not the actual surface temperature. The bolometric midnight temperature (\texttt{diviner\_tbol\_midnight}, range 90--188\,K) includes both rock and regolith contributions and is closer to actual surface temperature. Both channels are retained as independent thermal measurements with complementary information content.

\textbf{Clementine albedo units.} The USGS Clementine UVVIS 750\,nm mosaic stores photometrically corrected albedo as 8-bit digital numbers (DN, range 0--254) rather than calibrated reflectance. Cross-calibration against M\textsuperscript{3} 750\,nm yields $R^2 = 0.34$ with spatially varying ratios (0.0018--0.0032) due to fundamentally different photometric correction models (Lunar Lambert for Clementine vs. Hapke for M\textsuperscript{3}). Since all channels undergo z-score normalization, absolute units do not affect pretraining. Only spatial patterns matter.

\textbf{WAC Hapke longitude convention.} Four of eight WAC Hapke tiles used a 0--360$^\circ$ longitude convention while the remaining four used $\pm180^\circ$. Initial mosaicking without longitude wrapping yielded only 38.9\% coverage. Correcting the convention by wrapping tiles at 180$^\circ$ before mosaicking restored the expected 77.8\% coverage.

\textbf{LP GRS spatial resolution.} The Lunar Prospector Gamma-Ray Spectrometer provides elemental abundances (FeO, TiO\textsubscript{2}) at $\sim$60\,km effective spatial resolution from 11,306 point measurements. These are interpolated onto the 128 ppd grid ($\sim$237\,m/pixel) using radial basis function interpolation, producing smooth fields that vary at scales much coarser than the pixel grid.

\subsection{Normalization Statistics}

All channels are z-score normalized: $x_{\text{norm}}=(x-\mu)/\sigma$, where statistics are computed from 200 random $256\times256$ windows per channel. Missing data (NaN) is set to 0.0 after normalization. Tab.~\ref{tab:supp6} reports the complete statistics.

\section{Downstream Evaluation Details}

\subsection{Feature Extraction Protocol}

For all downstream tasks, the pretrained encoder is run with mask ratio 0 (all tokens visible), producing encoder outputs for each modality group.

\textbf{Classification and regression.} Tokens within each group are mean-pooled (excluding ghost tokens via the \texttt{has\_group} mask), then averaged across all available groups to produce a single 768-dimensional feature vector per patch:

\begin{equation}\label{eq:featpool}
\mathbf {f} = \frac {1}{|\mathcal {G}_{\text {valid}}|} \sum _{g \in \mathcal {G}_{\text {valid}}} \frac {1}{|T_g|} \sum _{i \in T_g} \mathbf {h}_i^g
\end{equation}

where $\mathcal{G}_{\text{valid}}$ is the set of groups with valid data and $T_g$ is the set of non-ghost tokens for group $g$.

\textbf{Segmentation.} Surface group tokens are used directly (256 tokens forming a $16\times16$ spatial grid) to preserve spatial information. The surface group has 100\% coverage, so all 256 tokens are always valid.

\subsection{Downstream Head Architectures}

\textbf{Linear probe.} A single linear layer: Linear(768, $C$) where $C$ is the number of classes or output dimensions.

\textbf{MLP head.} For classification and regression tasks: Linear(768, 384) $\to$ GELU $\to$ Dropout(0.1) $\to$ Linear(384, $C$).

\textbf{Segmentation head.} The $16\times16$ token grid is upsampled to $256\times256$ via two transposed convolutions:
\begin{align*}
&\text {ConvTranspose2d}(768, 128, k{=}4, s{=}4) \to \text {GELU} \\
&\to \text {ConvTranspose2d}(128, C, k{=}4, s{=}4)
\end{align*}
producing a $C$-channel prediction at the original input resolution ($256\times256$).

\textbf{Scratch baseline.} A lightweight CNN with three convolutional layers (28 $\to$ 64 $\to$ 128 $\to$ 256 channels, strides 4/2/2) with BatchNorm and ReLU, followed by adaptive average pooling and a linear classifier. This baseline processes all 28 raw channels directly.

\subsection{Training Hyperparameters}

Tab.~\ref{tab:supp7} provides the complete downstream training configuration. Fine-tuning uses a discriminative learning rate: the encoder's unfrozen blocks (layers 9--12) receive 0.01$\times$ the head learning rate. Tokenizers and positional embeddings remain frozen. Feature dropout (0.3) is applied to the extracted feature vector before the head during fine-tuning to prevent overfitting.

\subsection{Task Label Definitions}

\textbf{Geology classification (49 classes).} Each $256\times256$ patch is assigned the USGS Unified Geologic Map unit that covers the majority of pixels within the patch. The 49 classes span five stratigraphic periods and include mare basalts, highland materials, impact basin deposits, and crater ejecta units.

\textbf{Age classification (5 classes).} Geologic units are grouped into five stratigraphic age periods: pre-Nectarian ($>$3.92\,Ga), Nectarian (3.92--3.85\,Ga), Imbrian (3.85--3.2\,Ga), Eratosthenian (3.2--1.1\,Ga), and Copernican ($<$1.1\,Ga). The mapping is determined by the unit name prefix in the geologic map database.

\textbf{Composition regression.} Target values are the mean FeO and TiO\textsubscript{2} weight fractions within each patch, derived from LP GRS data. The metric is $R^2$ averaged over both elements. Since LP GRS has $\sim$60\,km effective resolution, this task evaluates whether the encoder captures regional geochemical trends rather than fine-grained spatial patterns.

\textbf{Cross-modal prediction.} The model predicts four thermal channels (\texttt{diviner\_tbol\_midnight}, \texttt{diviner\_temp\_night}, \texttt{rock\_abundance}, \texttt{christiansen\_feature}) from non-thermal modality groups only. At inference, the thermal group's \texttt{has\_group} flag is set to zero, forcing the model to rely entirely on cross-modal information.

\textbf{Mare segmentation.} Binary pixel-level classification of mare basalts vs. highland terrain, derived from the USGS Unified Geologic Map. Mare units are identified by their compositional classification in the map database.

\textbf{Crater segmentation.} Binary pixel-level segmentation of large impact craters ($>$10\,km diameter) from the Robbins global crater catalog. Crater interiors (within the rim) are labeled as positive.

\subsection{Dataset Split Statistics}

The 16,200 total patches (from a $180\times90$ non-overlapping grid) are split 70/15/15 into train (11,340), validation (2,430), and test (2,430) sets using a fixed random seed. Split assignments are stored in HDF5 metadata for exact reproducibility. For geographic split evaluation, patches are assigned by latitude band: training on $\pm60^\circ$, validation on 60--70$^\circ$ (both hemispheres), and testing on 70--80$^\circ$ (both hemispheres). This simulates deployment to under-explored polar regions with different geology, illumination conditions, and modality coverage compared to equatorial training data.

\section{Pretraining Details}

\subsection{Complete Hyperparameter Table}

Tab.~\ref{tab:supp8} provides the full pretraining configuration.

\textbf{Mixed precision.} We use bfloat16 rather than float16 for mixed-precision training. bfloat16 has the same exponent range as float32 (8 exponent bits vs. float16's 5), so no loss scaling or GradScaler is needed. The contrastive loss computation is explicitly cast to float32 to avoid numerical instability in the softmax temperature scaling.

\subsection{Training Dynamics}

Training proceeds smoothly over 100 epochs. Key observations:
\begin{itemize}
\item \textbf{Per-group convergence rates.} Gravity (MSE 0.003) converges within $\sim$20 epochs due to its smooth, low-frequency spatial structure. Radar remains the hardest group throughout training (final MSE 0.262), reflecting its sparse, high-frequency texture and limited 16--18\% coverage. Surface and Hapke groups achieve similar final MSE (0.077), while thermal (0.136) and spectral (0.175) settle at intermediate values.
\item \textbf{Contrastive loss.} $\mathcal{L}_{\text{NCE}}$ stabilizes by epoch $\sim$30, indicating that per-group encoder representations achieve modality-invariant alignment early in training.
\item \textbf{Proportional weighting dynamics.} The self-balancing loss weights shift over training: gravity's weight decreases as its loss drops, while radar's weight increases, automatically focusing learning capacity on harder groups.
\item \textbf{Validation.} Validation loss (computed on 2,000 random crops from the global map) closely tracks training loss with no evidence of overfitting, consistent with the unlimited random crop sampling strategy that ensures every training sample is effectively unique.
\end{itemize}

\section{Reproducibility and Data Release}

\subsection{Compute Requirements}

\textbf{Storage.} The complete dataset requires approximately: 72\,GB for aligned GeoTIFFs (source of truth), 115\,GB for pre-normalized memory-mapped arrays (training), and 57\,GB for the HDF5 evaluation dataset. Raw M\textsuperscript{3} data (used only during pipeline construction) requires an additional $\sim$943\,GB.

\subsection{Software Stack}

The implementation uses:
\begin{itemize}
\item PyTorch 2.7 with CUDA, Python 3.11
\item NumPy, rasterio, GDAL (geospatial I/O)
\item h5py (HDF5 dataset construction)
\item SciPy (LP GRS interpolation)
\end{itemize}
No custom CUDA kernels are required. Self-attention uses PyTorch's built-in \texttt{scaled\_dot\_product\_attention}, which automatically dispatches to FlashAttention-2 when available on compatible hardware.

\subsection{Data and Code Release}

We publicly release Moonstone in two locations:
\begin{itemize}
\item \textbf{Data (HuggingFace):} \url{https://huggingface.co/datasets/ayushprd/Moonstone} hosts the pre-normalized 28-channel memory-mapped arrays ($\sim$115\,GB), channel normalization statistics, train/val/test split metadata, aligned GeoTIFFs ($\sim$72\,GB), and the pretrained MG-MAE checkpoint.
\item \textbf{Code (GitHub):} \url{https://github.com/ayushprd/Moonstone} provides the complete pretraining and benchmark code, including the full 15-step data-processing pipeline, model architecture, training scripts, downstream evaluation for all six tasks, and baseline implementations.
\end{itemize}
All source data is derived from publicly accessible NASA PDS, USGS, and ISRO archives. No proprietary data is used. The complete processing pipeline (steps 1--15) is included for full reproducibility from raw downloads.

\subsection{Reproducibility Notes}

\textbf{DDP determinism.} All stochastic masking decisions (complementary masking, anchor group selection) are generated on rank 0 and broadcast to all ranks via \texttt{torch.distributed.broadcast}, ensuring identical mask patterns across GPUs. All random operations use PyTorch tensor operations rather than Python's random module for DDP compatibility.

\textbf{Fixed splits.} Train/val/test split assignments are stored in HDF5 metadata with a fixed random seed, ensuring exact reproducibility of all downstream evaluations.

\textbf{Few-shot evaluation.} Each few-shot experiment is repeated 5 times with seeds 0--4, sampling different support sets. Results are reported as mean $\pm$ standard deviation.

\section{Design Evolution}

The final MG-MAE architecture emerged through systematic exploration of design alternatives. We document these design decisions and the reasoning behind them, as they provide insight into effective strategies for multi-modal planetary foundation models.

\subsection{Per-Channel vs. Grouped Tokenization}

Our initial design assigned each of the 28 channels its own Conv2d(1, $D$, $P$) tokenizer, yielding 28 independent projection layers with a total of 181.4M parameters ($\sim$90.8M encoder, $\sim$90.6M decoder with 28 independent decoders).

This per-channel approach discards intra-group correlations. For example, the eight M\textsuperscript{3} spectral bands sample a continuous reflectance spectrum, but per-channel tokenizers process each band independently, losing the spectral shape information that is diagnostic of mineral composition. Similarly, the four Hapke reflectance bands encode a smooth spectral curve whose shape indicates surface maturity.

Grouped tokenization replaces 28 per-channel tokenizers with 7 per-group multi-channel tokenizers, reducing parameters by 44\% (181.4M $\to$ 101.6M) while capturing intra-group correlations. The ablation study in the main paper (Table 6) confirms a +3.3\% geology accuracy gain from grouped tokenization.

\subsection{Dirichlet vs. Complementary Masking}

The per-channel architecture used Dirichlet($\alpha$=1) sampling to allocate a token budget across available channels: for each sample, a random fraction was drawn from a Dirichlet distribution to determine how many tokens each channel retained. While mathematically elegant, this approach had two drawbacks:
\begin{enumerate}
\item \textbf{Complexity without benefit.} The Dirichlet distribution adds implementation complexity but showed no clear improvement over uniform random masking in preliminary experiments.
\item \textbf{No cross-modal forcing.} Dirichlet sampling does not systematically create conditions where the model must reconstruct one modality from another, which is a key learning objective for cross-modal representations.
\end{enumerate}
Complementary masking (Sec. 1.5) replaces this with a simpler, more targeted mechanism: with probability 0.5, designated anchor groups retain 100\% of tokens while others mask at 90\%. This directly forces cross-modal reconstruction and yields +1.4\% geology accuracy (main paper, Table 6).

\subsection{Independent vs. Shared Decoder}

The per-channel architecture used 28 independent decoders, one per channel, with no information exchange during reconstruction. This design prevents the decoder from using cross-modal correlations. For example, it cannot use surface morphology to improve thermal reconstruction.

The shared decoder in MG-MAE (4 layers, 384-d, 6 heads) processes all modality groups with a single set of transformer blocks, preceded by a cross-modal attention layer that allows each group's decoder tokens to attend to encoder tokens from all groups. This cross-modal decoder attention contributes +1.8\% geology accuracy and +1.7 mIoU on craters (main paper, Table 6).

The cross-modal prediction task provides direct evidence: the model achieves $R^2 = 0.987$ when reconstructing thermal channels from non-thermal inputs, demonstrating that learned cross-modal correlations transfer to held-out data.

\subsection{Earth--Moon Domain Gap}

A natural question is whether representations from Earth observation (EO) foundation models can transfer to lunar data. Using learned linear adapters to project 28 lunar channels to each model's expected input dimensionality while preserving pretrained patch embeddings, we evaluated SatMAE, Prithvi-EO-2.0, and TerraMind on downstream tasks.

The best EO model (Prithvi-EO-2.0) improved geology accuracy by only +1.9\% over training from scratch, compared to MG-MAE's +16.2\% improvement. This result validates the core premise of our work. EO models learn features specific to Earth (vegetation indices, ocean color, atmospheric scattering, seasonal phenology), none of which have lunar counterparts. The fundamental domain gap between terrestrial and planetary imagery requires domain-specific pretraining, even when adapter protocols provide compatible input distributions.

\section{Extended Visualizations}

\begin{figure}[t]
\centering
\includegraphics[width=0.85\textwidth]{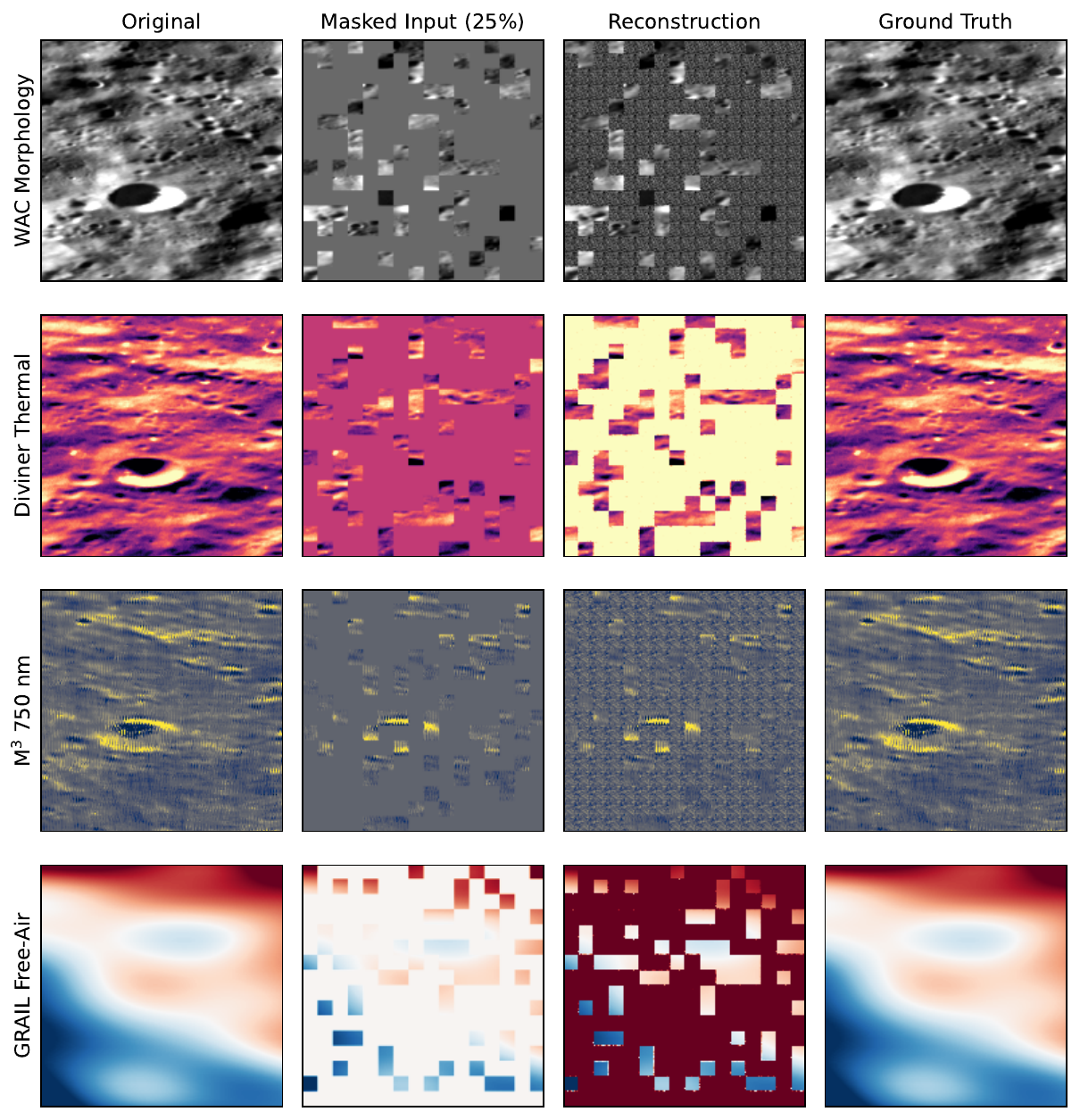}
\caption{Additional MG-MAE reconstruction examples across modality groups and lunar terrains. From left to right: original patch, masked input (25\% visible tokens, gray = masked), model reconstruction, and ground truth.}
\label{fig:supp1}
\end{figure}

Fig.~\ref{fig:supp1} provides additional reconstruction examples across diverse lunar terrains beyond those shown in the main paper (Fig.~2). The model reconstructs coherent spatial patterns from only 25\% visible tokens across physically distinct modalities. Reconstruction quality is highest for gravity (smooth, low-frequency fields) and surface morphology (strong spatial autocorrelation), and lowest for radar (sparse, high-frequency texture with only 16--18\% coverage). The model captures both large-scale structure (mare boundaries, crater rims) and fine-scale texture (spectral variations, thermal gradients).

\begin{table}[t]
\centering
\caption{Parameter count breakdown for MG-MAE (ViT-Base configuration).}
\label{tab:supp1}
\begin{tabular}{llr}
\toprule
Component & Details & Parameters \\
\midrule
\multicolumn{3}{l}{\textit{Per-group tokenizers} (\texttt{Conv2d($n_g$, 768, 16, 16)})} \\
\quad Surface ($n_g{=}4$)     & $4\times768\times256 + 768$ & 786,432 \\
\quad Thermal ($n_g{=}4$)     &                             & 786,432 \\
\quad Spectral ($n_g{=}8$)    & $8\times768\times256 + 768$ & 1,572,864 \\
\quad Gravity ($n_g{=}3$)     & $3\times768\times256 + 768$ & 589,824 \\
\quad Radar ($n_g{=}2$)       & $2\times768\times256 + 768$ & 393,216 \\
\quad Hapke ($n_g{=}4$)       &                             & 786,432 \\
\quad Composition ($n_g{=}3$) &                             & 589,824 \\
Tokenizer subtotal            &                             & 5,505,024 \\
\midrule
\multicolumn{3}{l}{\textit{Shared encoder} (12 transformer blocks)} \\
\quad Self-attention (per block) & QKV: $768{\to}2304$, proj: $768{\to}768$ & 2,362,368 \\
\quad MLP (per block)            & $768{\to}3072{\to}768$   & 4,722,432 \\
\quad LayerNorms (per block)     & $2\times768\times2$      & 3,072 \\
\quad Per block total            &                          & 7,087,872 \\
\quad $\times12$ blocks          &                          & 85,054,464 \\
\quad Encoder LayerNorm          &                          & 1,536 \\
Encoder subtotal (incl.\ embeddings) &                      & $\sim$85,600,000 \\
\midrule
\multicolumn{3}{l}{\textit{Decoder}} \\
\quad Encoder-to-decoder projection & $768\to384$          & 295,296 \\
\quad Cross-modal attention         & MHA(384, 6 heads)    & 592,128 \\
\quad Self-attention blocks ($\times4$) & 384-d transformer blocks & 4,740,096 \\
\quad Decoder LayerNorm             &                      & \\
\multicolumn{3}{l}{\textit{Per-group prediction heads} (\texttt{Linear($384$, $n_g\times16^2$)})} \\
\quad 7 heads (varying $n_g$)       &                      & $\sim$2,800,000 \\
\midrule
\multicolumn{3}{l}{\textit{Embeddings}} \\
\quad CLS token                     & $1\times768$         & \\
\quad 7 type embeddings             & $7\times768$         & 5,376 \\
\quad Positional embeddings         & $257\times768$ (sinusoidal, frozen) & 0 \\
\quad Mask tokens (decoder)         & $7\times384$         & 2,688 \\
\midrule
Total                               &                      & $\sim$101.6M \\
\bottomrule
\end{tabular}
\end{table}

\begin{table}[t]
\centering
\caption{Coverage-adaptive mask ratios. Groups with lower coverage retain more
visible tokens.}
\label{tab:supp2}
\begin{tabular}{lcccccccc}
\toprule
 & Surface & Thermal & Spectral & Gravity & Radar & Hapke & Comp. & Mean \\
\midrule
Coverage $c_g$   & 1.000 & 0.710 & 0.669 & 1.000 & 0.174 & 0.778 & 0.993 & 0.761 \\
Mask ratio $m_g$ & 0.786 & 0.742 & 0.736 & 0.786 & 0.662 & 0.753 & 0.785 & --- \\
Visible tokens   & $\sim$55 & $\sim$66 & $\sim$68 & $\sim$55 & $\sim$87 & $\sim$63 & $\sim$55 & --- \\
\bottomrule
\end{tabular}
\end{table}

\begin{table}[t]
\centering
\caption{Data sources for the 28-channel Moonstone dataset. All data are publicly
available from the indicated archives.}
\label{tab:supp3}
\setlength{\tabcolsep}{4pt}
\begin{tabular}{lllllr}
\toprule
Instrument & Mission & Archive & Format & Native Res. & Size \\
\midrule
WAC morphology     & LRO           & USGS        & GeoTIFF        & 100 m/px   & $\sim$6 GB \\
LOLA + SLDEM2015   & LRO           & USGS/MIT    & GeoTIFF/IMG    & 118 m/px   & $\sim$8 GB \\
Diviner GDR L3     & LRO           & PDS Geosci. & IMG+LBL        & 128 ppd    & $\sim$6 GB \\
Diviner GHRM       & LRO           & PDS Geosci. & IMG+LBL        & 128 ppd    & $\sim$3 GB \\
M\textsuperscript{3} L2 REFIMG & Chandrayaan-1 & PDS ODE & ENVI BIL (f32) & 140--280 m & $\sim$943 GB \\
GRAIL gridded      & GRAIL         & PDS         & ASCII (2 ppd)  & $\sim$15 km & $\sim$100 MB \\
Mini-RF SAR        & LRO           & PDS         & GeoTIFF        & $\sim$150 m & $\sim$4 GB \\
WAC Hapke          & LRO           & PDS         & GeoTIFF        & 128 ppd    & $\sim$8 GB \\
Clem. UVVIS 750 nm & Clementine    & USGS        & GeoTIFF (u8)   & 256 ppd    & $\sim$2 GB \\
LP GRS elements    & Lunar Prosp.  & PDS         & ASCII table    & $\sim$60 km & $\sim$1 MB \\
\bottomrule
\end{tabular}
\end{table}

\begin{table}[t]
\centering
\caption{15-step data processing pipeline. All scripts are included in the released code.}
\label{tab:supp4}
\setlength{\tabcolsep}{4pt}
\begin{tabular}{clll}
\toprule
Step & Script & Description & Output \\
\midrule
1  & \texttt{step01\_download}    & Download WAC, LOLA, SLDEM, Diviner      & Raw products \\
2  & \texttt{step02\_align}       & Reproject to 128 ppd equirectangular    & Aligned GeoTIFFs \\
3  & \texttt{step03\_derive}      & Compute slope and roughness from DEM    & Derived layers \\
4  & \texttt{step04\_tile}        & Tile into $256\times256$ patches        & Initial HDF5 \\
5  & \texttt{step05\_geology}     & Add USGS geology labels                 & Labeled patches \\
6  & \texttt{step06\_sample}      & Create train/val/test splits (70/15/15) & Split metadata \\
7  & \texttt{step07\_validate}    & Validate HDF5 data quality              & Quality report \\
8  & \texttt{step08\_ode\_query}  & Query PDS ODE API for M\textsuperscript{3} products & Product list (887) \\
9  & \texttt{step09\_m3\_download} & Download M\textsuperscript{3} L2 reflectance products & ENVI cubes \\
10 & \texttt{step10\_m3\_mosaic}  & Mosaic M\textsuperscript{3} strips $\to$ 8 bands + 4 geometry & Global GeoTIFFs \\
11 & \texttt{step11\_m3\_register} & Cross-register M\textsuperscript{3} vs.\ WAC & Registration QA \\
12 & \texttt{step12\_m3\_integrate} & Build V2 HDF5 (16 channels)           & V2 dataset \\
13 & \texttt{step13\_align\_new}  & Download + align GRAIL, Mini-RF, etc.   & 12 new GeoTIFFs \\
14 & \texttt{step14\_build\_v4}   & Build V4 HDF5 (28 channels)             & V4 dataset \\
15 & \texttt{step15\_build\_mmap} & Convert to normalized mmap arrays       & Training data \\
\bottomrule
\end{tabular}
\end{table}

\begin{table}[t]
\centering
\caption{Selected M\textsuperscript{3} spectral bands and their scientific relevance.}
\label{tab:supp5}
\begin{tabular}{ccl}
\toprule
Band Index & Center $\lambda$ (nm) & Scientific Target \\
\midrule
6  & 750  & Continuum albedo reference \\
16 & 950  & Pyroxene/olivine 1\,$\mu$m absorption \\
19 & 1000 & 1\,$\mu$m absorption shoulder \\
31 & 1250 & Plagioclase feldspar \\
48 & 1580 & SWIR continuum \\
69 & 2000 & Pyroxene 2\,$\mu$m absorption \\
79 & 2817 & OH/H\textsubscript{2}O \\
81 & 2857 & OH/H\textsubscript{2}O absorption shoulder \\
\bottomrule
\end{tabular}
\end{table}

\begin{table}[t]
\centering
\caption{Per-channel normalization statistics. Mean and standard deviation computed
from 200 random $256\times256$ windows per channel GeoTIFF.}
\label{tab:supp6}
\setlength{\tabcolsep}{5pt}
\begin{tabular}{llrrl}
\toprule
Group & Channel & Mean & Std & Units \\
\midrule
\multirow{4}{*}{Surface} & \texttt{wac\_morphology} & 63.94 & 41.59 & DN \\
 & \texttt{elevation} & $-456.83$ & 2291.08 & m \\
 & \texttt{slope} & 7.70 & 6.55 & degrees \\
 & \texttt{roughness} & 37.07 & 34.59 & std(m) \\
\midrule
\multirow{4}{*}{Thermal} & \texttt{diviner\_tbol\_midnight} & 95.43 & 6.59 & K \\
 & \texttt{diviner\_temp\_night} & 4.68 & 2.43 & K \\
 & \texttt{rock\_abundance} & 0.0033 & 0.0051 & fraction \\
 & \texttt{christiansen\_feature} & 8.17 & 0.10 & $\mu$m \\
\midrule
\multirow{8}{*}{Spectral} & \texttt{m3\_750} & 0.0799 & 0.0261 & reflectance \\
 & \texttt{m3\_950} & 0.0930 & 0.0319 & reflectance \\
 & \texttt{m3\_1000} & 0.0959 & 0.0331 & reflectance \\
 & \texttt{m3\_1250} & 0.1166 & 0.0379 & reflectance \\
 & \texttt{m3\_1580} & 0.1354 & 0.0430 & reflectance \\
 & \texttt{m3\_2000} & 0.1594 & 0.0495 & reflectance \\
 & \texttt{m3\_2817} & 0.2071 & 0.0585 & reflectance \\
 & \texttt{m3\_2857} & 0.2138 & 0.0600 & reflectance \\
\midrule
\multirow{3}{*}{Gravity} & \texttt{grail\_freeair} & $-4.56$ & 128.31 & mGal \\
 & \texttt{grail\_bouguer} & 21.08 & 214.70 & mGal \\
 & \texttt{grail\_uncertainty} & 1.91 & 1.60 & mGal \\
\midrule
\multirow{2}{*}{Radar} & \texttt{minirf\_cpr} & 0.101 & 0.479 & log(1+ratio) \\
 & \texttt{minirf\_s1} & 0.958 & 2.793 & log(1+DN) \\
\midrule
\multirow{4}{*}{Hapke} & \texttt{wac\_hapke\_415nm} & 0.0254 & 0.0070 & reflectance \\
 & \texttt{wac\_hapke\_566nm} & 0.0368 & 0.0098 & reflectance \\
 & \texttt{wac\_hapke\_604nm} & 0.0397 & 0.0104 & reflectance \\
 & \texttt{wac\_hapke\_689nm} & 0.0454 & 0.0117 & reflectance \\
\midrule
\multirow{3}{*}{Composition} & \texttt{clementine\_uvvis\_750nm} & 41.55 & 23.02 & DN \\
 & \texttt{lpgrs\_tio2} & 0.0086 & 0.0118 & wt fraction \\
 & \texttt{lpgrs\_feo} & 0.0693 & 0.0461 & wt fraction \\
\bottomrule
\end{tabular}
\end{table}

\begin{table}[t]
\centering
\caption{Downstream training hyperparameters across evaluation modes.}
\label{tab:supp7}
\begin{tabular}{lccc}
\toprule
Parameter & Linear Probe & Fine-tune & Scratch \\
\midrule
Optimizer               & AdamW & AdamW & AdamW \\
Head learning rate      & $10^{-3}$ & $10^{-3}$ & $10^{-3}$ \\
Encoder learning rate   & --- (frozen) & $10^{-5}$ ($0.01\times$) & $10^{-3}$ \\
Weight decay            & 0.05 & 0.05 & 0.05 \\
Max epochs              & 50 & 50 & 50 \\
Early stopping patience & 5 & 5 & 5 \\
Label smoothing         & 0.1 & 0.1 & 0.1 \\
Feature dropout         & --- & 0.3 & --- \\
Batch size (cls./reg.)  & 32 & 32 & 32 \\
Batch size (segmentation) & 16 & 16 & 16 \\
LR schedule             & cosine + warmup & cosine + warmup & cosine + warmup \\
Warmup                  & min(2 ep, 10\%) & min(2 ep, 10\%) & min(2 ep, 10\%) \\
Gradient clipping       & 1.0 & 1.0 & 1.0 \\
Frozen encoder blocks   & all 12 & first 8 of 12 & --- \\
\bottomrule
\end{tabular}
\end{table}

\begin{table}[t]
\centering
\caption{Complete MG-MAE pretraining hyperparameters.}
\label{tab:supp8}
\begin{tabular}{ll}
\toprule
Parameter & Value \\
\midrule
\multicolumn{2}{l}{\textit{Architecture}} \\
\quad Encoder          & ViT-Base (12 layers, 768-d, 12 heads) \\
\quad Decoder          & 4 layers, 384-d, 6 heads \\
\quad Patch size       & $16\times16$ pixels \\
\quad Input size       & $256\times256$ pixels \\
\quad Total parameters & $\sim$101.6M \\
\midrule
\multicolumn{2}{l}{\textit{Optimization}} \\
\quad Optimizer        & AdamW \\
\quad $(\beta_1, \beta_2)$ & (0.9, 0.95) \\
\quad Weight decay     & 0.05 \\
\quad Learning rate    & $1.5\times10^{-4}$ \\
\quad LR schedule      & Linear warmup $\to$ cosine decay \\
\quad Warmup epochs    & 10 \\
\quad Minimum LR       & $1\times10^{-6}$ \\
\quad Gradient clipping & \texttt{max\_norm} = 1.0 \\
\midrule
\multicolumn{2}{l}{\textit{Training}} \\
\quad Epochs                & 100 \\
\quad Batch size (per GPU)  & 64 \\
\quad GPUs                  & 2$\times$ NVIDIA H100 80 GB \\
\quad Gradient accumulation & 2 steps \\
\quad Effective batch size  & 256 \\
\quad Mixed precision       & bfloat16 \\
\quad Training time         & $\sim$20 hours ($\sim$40 GPU-hours) \\
\quad Throughput            & $\sim$270 samples/s \\
\midrule
\multicolumn{2}{l}{\textit{Masking}} \\
\quad Standard mask ratio                     & 0.75 \\
\quad Complementary mask ratio (non-anchor)   & 0.90 \\
\quad Complementary masking probability       & 0.5 \\
\quad Coverage-adaptive $\alpha$              & 0.15 \\
\midrule
\multicolumn{2}{l}{\textit{Loss}} \\
\quad InfoNCE temperature $\tau$    & 0.07 \\
\quad InfoNCE weight $\lambda$      & 0.1 \\
\quad Spectral continuity weight $\mu$ & 0.01 \\
\quad Reconstruction loss          & MSE (masked patches only) \\
\quad Loss group weighting         & Proportional (self-balancing) \\
\midrule
\multicolumn{2}{l}{\textit{Validation}} \\
\quad Validation frequency & Every 5 epochs \\
\quad Validation samples   & 2,000 random crops \\
\bottomrule
\end{tabular}
\end{table}

\begin{table}[t]
\centering
\caption{Compute requirements for reproducing Moonstone.}
\label{tab:supp9}
\begin{tabular}{lll}
\toprule
Stage & Hardware & Time \\
\midrule
Data pipeline (steps 1--15)              & CPU + 1 TB storage & $\sim$3 days \\
MG-MAE pretraining (100 epochs)          & 2$\times$ H100 80 GB & $\sim$20 hours \\
Full downstream eval (6 tasks $\times$ 3 modes) & 1$\times$ GPU & $\sim$12--18 hours \\
\bottomrule
\end{tabular}
\end{table}

\end{document}